\documentclass{article}
\usepackage{ijcai26}

\usepackage{times}
\usepackage{soul}
\usepackage[utf8]{inputenc}
\usepackage[small]{caption}
\usepackage{graphicx}
\usepackage{amsmath,amssymb}
\usepackage{amsthm}
\usepackage{booktabs}
\usepackage{algorithm}
\usepackage{algpseudocode}
\usepackage[switch]{lineno}
\usepackage{microtype}
\usepackage{mathtools}
\usepackage{enumitem}
\usepackage{array}
\usepackage{natbib}
\usepackage{placeins}
\usepackage{tikz}
\usepackage{url}
\usepackage[hidelinks]{hyperref}
\usetikzlibrary{arrows.meta,positioning}

\title{From Monolithic to Modular: Segment-level Automatic Prompt
Optimization}

\author{
Nikita Kulin$^1$
\and
Viktor Zhuravlev$^2$\and
Artur Khairullin$^3$\and
Sergey Muravyov$^4$\and
Ilya Makarov$^5$\and
Daniil Sukhorukov$^6$\And
Ekaterina Averkova$^7$\\
\affiliations
$^{1,2,3,4,7}$ITMO University\\
$^{5,6}$AXXX\\
\emails
nikita.kulin@itmo.ru$^1$,
vnzhuravlev@itmo.ru$^2$,
arkhairullin@itmo.ru$^3$,
smuravyov@itmo.ru$^4$,
iamakarov@hse.ru$^5$,
d.sukhorukov@axxx.tech$^6$,
ekaterina.averkova@itmo.ru$^7$
}

\begin{document}

\maketitle

\begin{abstract}
    Automatic Prompt Optimization (APO) often rewrites prompts monolithically, which can improve
one behavior while degrading others. We present SAPO, a segment-level APO method that decomposes
prompts into role, context, tasks, and output format, then applies targeted improvements based on top-5
and bottom-5 examples. The optimization loop uses one LLM with static meta-prompts and structured
outputs for segmentation, weakness analysis, and candidate generation. We describe a train/validation
protocol and a two-stage generation process: (1) segment-level diagnosis and recommendation extraction,
(2) candidate synthesis constrained by weak/strong segment signals. Using the evaluation setup across
SQuADv2, TweetEval, XSUM, CommonGen, and GSM8K on GPT-3.5-Turbo and GPT-4o-mini, SAPO achieves the best average score against Zero-shot and strong APO baselines including APE,
OPRO, EvoPrompt, GEPA, and StraGO.
\end{abstract}

\section{Introduction}
Large language models (LLMs) are increasingly used as general-purpose interfaces for NLP tasks, including instruction following, reasoning, and generation \citep{gpt3,cot,instructgpt}. In many practical settings, adaptation has shifted from finetuning toward prompt design, where system behavior is controlled through natural-language instructions \citep{ape_iclr2023}. As a consequence, prompt quality becomes a primary reliability bottleneck: small wording changes can produce large behavioral shifts, while manual prompt iteration remains costly and unstable \citep{ape_iclr2023,protegri}.

Automatic Prompt Optimization (APO) addresses this problem by iteratively generating and selecting improved prompts from data and model feedback \citep{ape_iclr2023,protegri,opro,evoprompt,gepa_iclr2026}. Such approaches are particularly attractive because they are deployment-friendly and compatible with black-box APIs.

A persistent limitation, however, is that many APO pipelines still optimize prompts as monolithic strings. In practice, prompts are compositional artifacts: role framing, context grounding, task directives, and output-format constraints contribute differently to downstream behavior. Prior work reports prompt drifting and instability in iterative optimization, where edits that fix one subset of cases may degrade previously correct behavior \citep{strago}. Additional studies show that optimizer effectiveness is sensitive to model capability and setup choices \citep{revisiting_opro,llms_good_prompt_optimizers}, and that gains can be regime-dependent in compound systems \citep{prompt_optimization_coin_flip}. Thus, the core challenge is not merely generating more prompt variants, but controlling \emph{where} and \emph{how} edits are applied.

This paper presents Segment-level APO (SAPO), a modular optimization strategy over explicit prompt segments. Instead of one global rewrite, the method diagnoses weak and strong segments from ranked evidence and synthesizes candidates that preserve strengths while fixing weaknesses. Here, \textit{strong segments} are segments associated with correct model behavior on top-ranked examples, while \textit{weak segments} are segments associated with incorrect behavior on bottom-ranked examples.

\begin{figure}[h]
\centering
\includegraphics[width=1.0\linewidth]{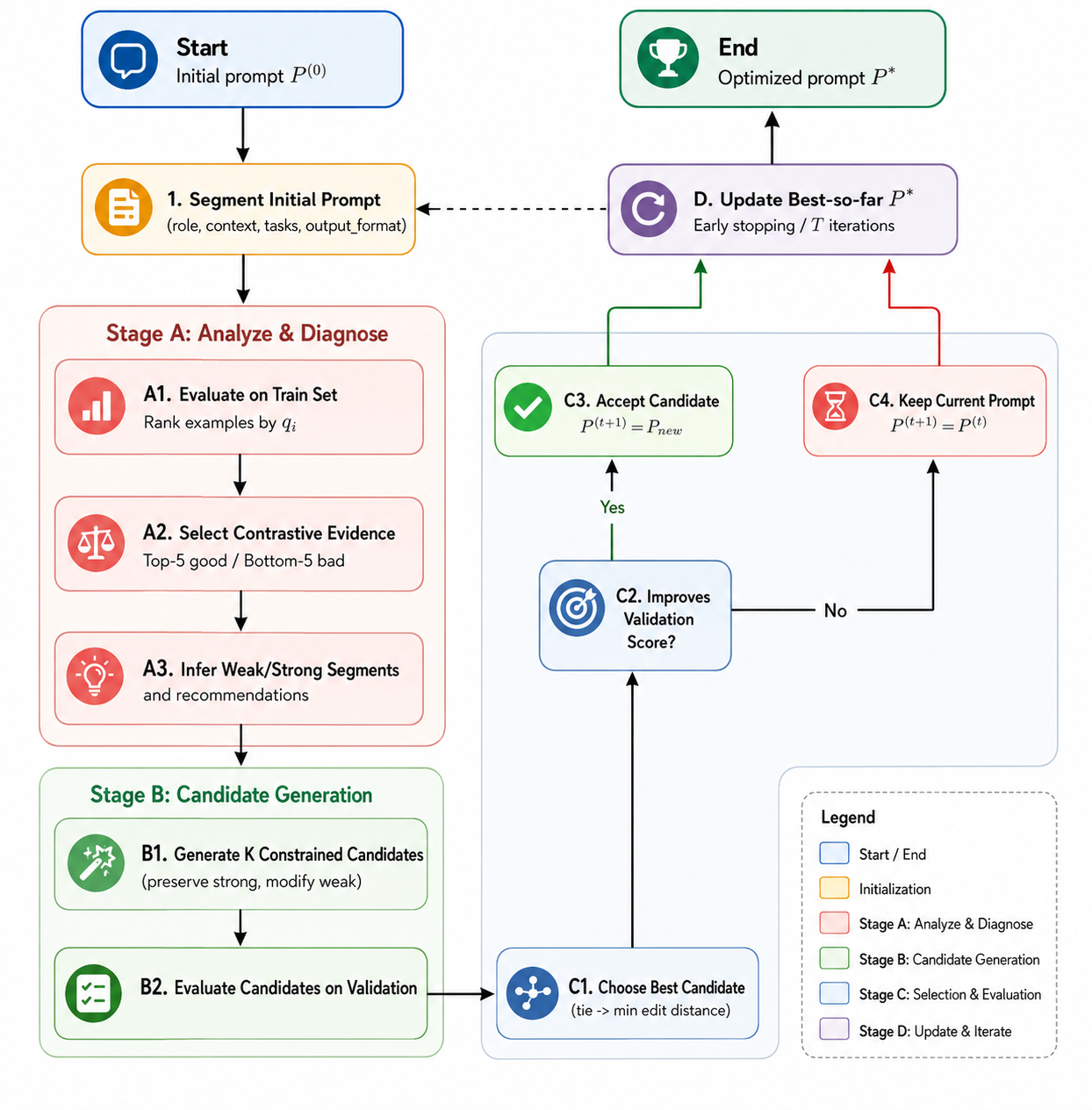}
\caption{SAPO loop. Starting from the initial prompt $P^{(0)}$, the method first segments instructions into role/context/tasks/output format, then performs contrastive evidence extraction on train data to diagnose weak and strong segments. Next, it generates constrained candidates that preserve strong segments while revising weak ones, evaluates them on validation data, applies edit-distance tie-break under equal scores, and accepts an update only when validation performance improves; otherwise the current prompt is retained.}

\label{fig:sapo_overview}
\end{figure}

\textbf{Compared with prior autoprompting methods, SAPO contributes:}
\begin{enumerate}[leftmargin=*]
    \item \textbf{Segment-explicit optimization objective:} prompts are decomposed into role, context, tasks, and output format, enabling targeted rather than monolithic edits.
    \item \textbf{Constrained candidate synthesis and selection:} candidates preserve strong segments, revise weak segments, and use an edit-distance tie-break for conservative updates.
    \item \textbf{Contrastive diagnostic stage:} update decisions are grounded in top/bottom evidence, including class-aware handling for discrete per-example metrics.
\end{enumerate}

We evaluate SAPO on five datasets with different task types: SQuADv2, TweetEval, XSUM, CommonGen, and GSM8K, and report results on two model backbones: GPT-3.5-Turbo and GPT-4o-mini. SAPO achieves the best average score against APE, OPRO, EvoPrompt, GEPA, and StraGO. Relative to the strongest competing baseline by average score, gains are \textbf{+5.13\%} on GPT-3.5-Turbo and \textbf{+7.25\%} on GPT-4o-mini.

\section{Related Work}
\paragraph{Search-based APO methods.}
Early and widely used APO methods formulate prompt improvement as iterative search over textual candidates. APE optimizes instruction candidates via generation and selection \citep{ape_iclr2023}, while ProTeGi applies textual gradients with beam search and bandit-style selection \citep{protegri}. OPRO and EvoPrompt further extend search with trajectory-conditioned optimization and evolutionary operators, respectively \citep{opro,evoprompt}. GEPA introduces reflective prompt evolution, providing modular modifications through optimization \citep{gepa_iclr2026}. Despite their effectiveness, these methods primarily operate as global prompt-level rewrites rather than as explicit segment-constrained updates \citep{ape_iclr2023,opro,evoprompt,promptbreeder}.

\paragraph{Structured and modular optimization trends.}
In parallel, the field is moving toward structured optimization workflows. DSPy and TextGrad cast optimization in programmatic or graph-based forms rather than single prompt rewrites \citep{dspy,textgrad}. Promptomatix emphasizes modular orchestration and cost-aware refinement \citep{promptomatix}. More recently, section-local optimization explicitly operates over fixed prompt components and reports improved robustness in small-model settings \citep{modular_prompt_optimization}. Collectively, this line of work indicates that structural decomposition can improve interpretability and reduce destructive interference between edits.

\paragraph{Gap summary.}
Existing literature provides either strong search performance or improved robustness diagnostics \citep{opro,evoprompt,strago,modular_prompt_optimization,gepa_iclr2026}, but there is still limited evidence on simple black-box pipelines that jointly enforce segment-level controllability, preserve known-strong prompt components, and remain implementation-light under a unified multi-task protocol.

\paragraph{Positioning of SAPO.}
SAPO sits at the intersection of static meta-prompting and structure-aware optimization. It preserves the deployment simplicity of black-box APO, while introducing explicit segment-level diagnosis and constrained updates that preserve strong components and localize revisions. 

\section{Method}
\subsection{Problem Formulation}
Figure~\ref{fig:sapo_overview} provides the high-level pipeline view, while Algorithm~\ref{alg:sapo} specifies the executable optimization procedure.

Let an initial prompt template be $P^{(0)}$ with placeholder \texttt{\{input\}}. Let datasets be split into training and validation parts:
\begin{equation}
\mathcal{D}_{train}=\{(x_i,y_i)\}_{i=1}^{N_{tr}}, \quad
\mathcal{D}_{val}=\{(x_j,y_j)\}_{j=1}^{N_{val}}.
\end{equation}

Given one LLM $\mathcal{M}$ and task-dependent metric $Q_{\tau}$, the objective is:
\begin{equation}
P^* = \arg\max_{P \in \mathcal{P}} Q_{\tau}(P;\mathcal{D}_{val},\mathcal{M}).
\end{equation}

At iteration $t$, the optimizer applies an update operator $\mathcal{U}$ to the current prompt:
\begin{equation}
P^{(t+1)} = \mathcal{U}\!\left(P^{(t)}, \mathcal{D}_{train}, \mathcal{D}_{val}\right).
\end{equation}
The operator is \textit{acceptance-constrained}: if the best candidate generated at iteration $t$ does not improve validation quality, the prompt is kept unchanged. This converts the procedure into a monotone, validation-gated search over prompt space, which directly targets robustness against destructive rewrites.

Prompt structure is represented as four explicit segments:
\begin{equation}
S(P)=\{s_{role}, s_{context}, s_{tasks}, s_{output\_format}\}.
\end{equation}

\subsection{Algorithm Overview}

The optimization loop can be summarized as:
\begin{enumerate}[leftmargin=*]
    \item Decompose the current prompt into segments \{role, context, tasks, output\_format\}.
    \item Run Stage A: evaluate on train, extract top-5/bottom-5 evidence, infer weak/strong segments and recommendations.
    \item Run Stage B: generate $K$ constrained prompt candidates.
    \item Evaluate candidates on validation; choose the best by score, with edit-distance tie-break.
    \item Accept the candidate only if validation score improves; otherwise keep the current prompt.
\end{enumerate}
Figure~\ref{fig:sapo_overview} and Algorithm~\ref{alg:sapo} formalize this loop.

\begin{algorithm}[h]
\caption{Segment-level APO (SAPO).}
\label{alg:sapo}
\begin{algorithmic}[1]
\Require Initial prompt $P$, train set $\mathcal{D}_{train}$, validation set $\mathcal{D}_{val}$, candidates $K$, iterations $T$
\State $P^* \gets P$; $Q^* \gets -\infty$
\For{$t=1$ to $T$}
    \State Segment current prompt into \{role, context, tasks, output\_format\}
    \State Stage A: evaluate $P$ on $\mathcal{D}_{train}$, build top-5/bottom-5 evidence, infer weak/strong segments and recommendations
    \State Stage B: generate $K$ improved candidates
    \State Evaluate candidates on $\mathcal{D}_{val}$ and choose best candidate $\tilde{P}$ (tie: minimum edit distance)
    \If{$Q_{\tau}(\tilde{P};\mathcal{D}_{val}) > Q_{\tau}(P;\mathcal{D}_{val})$}
        \State $P \gets \tilde{P}$
    \EndIf
    \If{$Q_{\tau}(P;\mathcal{D}_{val}) > Q^*$}
        \State $P^* \gets P$; $Q^* \gets Q_{\tau}(P;\mathcal{D}_{val})$
    \EndIf
\EndFor
\State \Return $P^*$
\end{algorithmic}
\end{algorithm}

\subsection{Prompt Segment Design}
SAPO optimizes four segments because they map to distinct and operationally separable control dimensions in instruction-based LLM use.

\paragraph{Role (\(s_{role}\)).}
This segment defines behavioral stance (e.g., classifier, summarizer, analyst). Prior prompt engineering literature shows that instruction framing can materially affect downstream behavior \citep{ape_iclr2023,protegri}.

\paragraph{Context (\(s_{context}\)).}
This segment encodes task grounding, input injection, and domain constraints (including \texttt{\{input\}} placement). It controls what information is available and how the model conditions on it.

\paragraph{Tasks (\(s_{tasks}\)).}
This segment specifies actionable requirements and decision rules. It is the main locus for correcting underspecified or ambiguous instructions.

\paragraph{Output format (\(s_{output\_format}\)).}
This segment governs the response schema and formatting constraints. It is critical in tasks where metric outcomes depend on strict label/output conventions (e.g., classification labels or short-form answers).

These four segments were selected because they provide a compact decomposition that is expressive enough for heterogeneous NLP tasks while remaining small enough for stable, low-cost iterative optimization. This design choice is aligned with broader prompt-structure taxonomies and recent trends toward structured and section-local prompt optimization \citep{pretrain_prompt_predict,prompt_report,dspy,textgrad,modular_prompt_optimization}.

\subsection{Two-stage Generation Pipeline}

\paragraph{Stage A: Evidence extraction and segment-level diagnosis.}
At iteration $t$, the current prompt $P^{(t)}$ is first evaluated on $\mathcal{D}_{train}$:
\begin{equation}
\hat{y}_i = \mathcal{M}(P^{(t)},x_i), \quad q_i=Q_{\tau}\!\left(P^{(t)}; (x_i,y_i), \mathcal{M}\right).
\end{equation}
Examples are ranked by $q_i$, and two contrastive evidence sets are extracted:
\begin{equation}
\mathcal{B}_{good}^{(t)} = \text{Top-5}(q_i), \quad
\mathcal{B}_{bad}^{(t)} = \text{Bottom-5}(q_i).
\end{equation}
For discrete per-example metrics (e.g., ExactMatch with $q_i \in \{0,1\}$), we use class-aware evidence selection: $\mathcal{B}_{good}^{(t)}$ is drawn first from positive examples ($q_i=1$), and $\mathcal{B}_{bad}^{(t)}$ is drawn first from negative examples ($q_i=0$). If one side has fewer than five examples, the remainder is backfilled from the global rank order.
Given $(\mathcal{B}_{good}^{(t)},\mathcal{B}_{bad}^{(t)})$ and segment decomposition $S(P^{(t)})$, one LLM with static meta-prompts infers structured diagnostic outputs:
\texttt{weak\_segments}, \texttt{strong\_segments}, and \texttt{recommendations}.
Intuitively, strong segments are associated with consistently successful evidence, while weak segments are associated with failure cases and become primary targets for revision.

\paragraph{Stage B: Candidate synthesis.}
The same LLM receives current prompt, segment decomposition, weak/strong labels, and recommendations, then generates $K$ improved prompt candidates $\{P_k^{(t)}\}_{k=1}^{K}$.
Candidate synthesis is explicitly constrained to preserve segments listed in \texttt{strong\_segments} and primarily modify segments listed in \texttt{weak\_segments}, reducing cross-segment interference.

Each candidate $P_k^{(t)}$ is evaluated on $\mathcal{D}_{val}$:
\begin{equation}
Q_{k,\tau}^{(t)}=Q_{\tau}(P_k^{(t)};\mathcal{D}_{val},\mathcal{M}).
\end{equation}
The best candidate is accepted only if it improves the current validation score. Under score ties, we select the candidate with the smallest edit distance to the current prompt:
\begin{equation}
\tilde{P}^{(t)}=\arg\max_{P_k^{(t)}} Q_{k,\tau}^{(t)}, \quad
\text{tie-break by } \min d_{\text{edit}}(P_k^{(t)}, P^{(t)}).
\end{equation}
This conservative tie-break favors minimal edits and helps preserve validated prompt behavior.

For compactness, Stages A--B define an update operator:
\begin{equation}
\tilde{P}^{(t)}=\mathcal{U}\!\left(P^{(t)},\mathcal{B}_{good}^{(t)},\mathcal{B}_{bad}^{(t)}\right),
\end{equation}
where $\mathcal{U}$ performs diagnosis, constrained candidate synthesis, and tie-aware selection.

An example of SAPO optimization trajectory is shown in Figure~\ref{fig:sapo_e5_temp}. Meta-prompts for each stage and method complexity analysis are provided in Appendix~\ref{appendix:prompt} and Appendix~\ref{app:costs}.

\section{Experimental Setup}
\paragraph{Datasets and task coverage.}
We evaluate on five datasets spanning extractive QA, social NLP classification, abstractive summarization, constrained commonsense generation, and mathematical reasoning:
SQuADv2 \citep{squad2}, TweetEval \citep{tweeteval}, XSUM \citep{xsum}, CommonGen (CG) \citep{commongen}, and GSM8K \citep{gsm8k}.
This mix is intended to stress different prompt components (role, task specification, response format, and domain context) under a shared APO loop. In our evaluation, we use BERTScore F1 \citep{bertscore} for (XSUM, CG, SQuAD2), F1 -- (TweetEval), and ExactMatch -- (GSM8K).

\paragraph{Models.}
We report results for GPT-3.5-Turbo and GPT-4o-mini to cover two commonly used API models with different capability/cost profiles.

\paragraph{Data split and protocol.}
Each dataset is split into train/validation/test with sizes 150/100/all. The train split is used for top/bottom evidence extraction, validation is used for candidate selection during optimization, and the held-out test split is used for final reporting. This separation is strictly enforced for all iterative methods to reduce selection leakage.

\paragraph{Generation settings and baselines.}
For candidate generation, we use $K=5$ candidates per iteration and a sampling temperature of $0$. These values are fixed across all experiments and are kept identical for SAPO and all iterative APO baselines for fairness. We compare against Zero-shot prompting (as a baseline and as an initial prompt for optimization) and representative APO methods: APE \citep{ape_iclr2023}, OPRO \citep{opro}, EvoPrompt \citep{evoprompt}, GEPA \citep{gepa_iclr2026}, and StraGO \citep{strago}.

\paragraph{Runs and reporting.}
Each method/model setup is run five times with different random seeds and data samples. Main tables report mean values across runs.
\FloatBarrier

\section{Results}
\label{sec:results}

Across both model settings, SAPO delivers the highest average score among compared baselines on the reported benchmark suite. Unless explicitly stated otherwise, all numbers in this section are evaluated on the held-out test split. The largest relative gains appear on GSM8K and XSUM, consistent with the value of segment-level control in settings that require strict answer extraction or tightly constrained generation. The SAPO-optimized prompts are provided in Appendix~\ref{appendix:prompt}.

On GPT-3.5-Turbo, SAPO achieves the highest average score. It yields gains on SQuAD2, GSM8K, CommonGen, and TweetEval, while trailing GEPA on XSUM. The largest relative gain appears on GSM8K, suggesting that segment-level control and task-focused refinement are particularly beneficial for exact-answer reasoning settings.

On GPT-4o-mini, SAPO again delivers the highest average score and outperforms all compared baselines on all five datasets. This pattern indicates that segment-level optimization is complementary to model capability scaling rather than specific to one backbone.

The average gain relative to the best competing baseline by average score is +5.13\% on GPT-3.5-Turbo and +7.25\% on GPT-4o-mini, indicating consistent benefits from constrained segment-local updates.

\begin{table}[h]
\centering
\footnotesize
\setlength{\tabcolsep}{1.2pt}
\caption{Test-split comparison on GPT-3.5-Turbo. Best values per column are in \textbf{bold}. ``Difference vs best baseline'' is computed per column against the strongest non-SAPO method in that column.}
\label{tab:gpt35}
\begin{tabular}{lcccccc}
\toprule
Method & SQuAD2 & GSM8K & CG & TweetEval & XSum & Avg \\
\midrule
Zero-shot & 0.812 & 0.527 & 0.824 & 0.483 & 0.801 & 0.689 \\
APE & 0.904 & 0.710 & 0.883 & 0.511 & 0.815 & 0.765 \\
OPRO & 0.871 & 0.715 & 0.873 & 0.536 & 0.828 & 0.765 \\
EvoPrompt & 0.895 & 0.697 & 0.889 & 0.521 & 0.837 & 0.768 \\
GEPA & 0.873 & 0.744 & 0.871 & 0.539 & \textbf{0.859} & 0.777 \\
StraGO & 0.915 & 0.717 & 0.908 & 0.523 & 0.834 & 0.779 \\
\textbf{SAPO} & \textbf{0.922} & \textbf{0.835} & \textbf{0.911} & \textbf{0.575} & 0.852 & \textbf{0.819} \\
\midrule
Diff. & +0.007 & +0.091 & +0.003 & +0.036 & -0.007 & +0.040 \\
Rel. diff. (\%) & +0.77 & +12.23 & +0.33 & +6.68 & -0.81 & +5.13 \\
\bottomrule
\end{tabular}
\end{table}

\begin{table}[h]
\centering
\footnotesize
\setlength{\tabcolsep}{1.2pt}
\caption{Test-split comparison on GPT-4o-mini. Best values per column are in \textbf{bold}. ``Difference vs best baseline'' is computed per column against the strongest non-SAPO method in that column.}
\label{tab:gpt4omini}
\begin{tabular}{lcccccc}
\toprule
Method & SQuAD2 & GSM8K & CG & TweetEval & XSum & Avg \\
\midrule
Zero-shot & 0.795 & 0.841 & 0.784 & 0.477 & 0.638 & 0.707 \\
APE & 0.851 & 0.884 & 0.823 & 0.551 & 0.690 & 0.760 \\
OPRO & 0.842 & 0.891 & 0.831 & 0.510 & 0.695 & 0.754 \\
EvoPrompt & 0.827 & 0.905 & 0.808 & 0.556 & 0.714 & 0.762 \\
GEPA & 0.923 & 0.880 & 0.823 & 0.539 & 0.695 & 0.772 \\
StraGO & 0.838 & 0.903 & 0.810 & 0.572 & 0.720 & 0.769 \\
\textbf{SAPO} & \textbf{0.931} & \textbf{0.952} & \textbf{0.875} & \textbf{0.593} & \textbf{0.789} & \textbf{0.828} \\
\midrule
Diff. & +0.008 & +0.047 & +0.044 & +0.021 & +0.069 & +0.056 \\
Rel. diff. (\%) & +0.87 & +5.19 & +5.29 & +3.67 & +9.58 & +7.25 \\
\bottomrule
\end{tabular}
\end{table}

\section{Discussion}
\paragraph{Why segment-level updates help.}
The observed gains are consistent with the hypothesis that explicit weak/strong segment separation reduces destructive interference between edits. Instead of globally rewriting prompts, SAPO localizes revisions to weak components while preserving high-performing structure, aligning with recent observations on prompt drifting in iterative APO \citep{strago}.

\paragraph{When improvements are likely.}
Improvements are most pronounced when tasks depend on strict output conventions or multi-part instructions. In such settings, preserving strong format/context segments appears as important as improving task wording; this is consistent with recent evidence that optimization success is regime-dependent and tied to task structure \citep{prompt_optimization_coin_flip}.

\paragraph{Limitations.}
A key limitation of this work is that SAPO uses a fixed and relatively small segment set. While this decomposition is practical and interpretable, it may not capture finer-grained prompt structure needed for some tasks. Future work should explore richer segment taxonomies and adaptive segment discovery, including LLM-based automatic generation of new segment types during optimization. This would allow the optimizer to expand or refine its own editing space dynamically, rather than relying on a predefined segmentation scheme. Meanwhile, our current setup uses fixed top-5/bottom-5 evidence windows, which may miss finer-grained error modes.

\section{Conclusion}
We presented SAPO, a modular APO framework that replaces monolithic rewrites with segment-level optimization. Under this protocol, SAPO achieves the strongest average performance against strong APO baselines and offers an interpretable pathway for future extensions in APO.

\section{Acknowledgments}
This work supported by the Ministry of Economic Development of the Russian Federation (IGK 000000C313925P4C0002), agreement No139-15-2025-010

\bibliographystyle{named}
\bibliography{ijcai26}

\appendix

\section{Ablation Study}
\label{sec:ablation}

\subsection{Segment Importance}
To quantify the contribution of individual prompt segments, we run a controlled ablation on all five datasets using the same task-specific metrics as in the main experiments. For each dataset, we treat the final SAPO prompt as the baseline and evaluate on test samples ablated variants relative to that baseline:
\begin{itemize}
    \item \(-\)Role: final prompt without role
    \item InitTasks: final prompt with tasks replaced by initial tasks
    \item InitCtx: final prompt with context replaced by initial context
    \item \(-\)RespFmt: final prompt without response format
\end{itemize}

\begin{table}[h]
\centering
\small
\setlength{\tabcolsep}{0.8pt}
\begin{tabular}{l c c c c c}
\toprule
Dataset & Baseline & $\Delta$(-Role) & $\Delta$(InitTasks) & $\Delta$(InitCtx) & $\Delta$(-RespFmt) \\
\midrule
SQuADv2 & - & +0.0210 & -0.0637 & +0.0000 & -0.0281 \\
TEval & - & -0.0744 & -0.0391 & -0.0160 & -0.0844  \\
XSUM & - & -0.0103 & -0.0811 & +0.0011 & -0.0328 \\
CG & - & -0.0016 & -0.0550 & -0.0005 & -0.0522 \\
GSM8K & - & -0.0801 & -0.2800 & +0.0000 & +0.0207  \\
\midrule
Average & - & \textbf{-0.0291} & \textbf{-0.1038} & \textbf{-0.0031} & \textbf{-0.0354} \\
\bottomrule
\end{tabular}
\caption{Segment ablation on five datasets. Each $\Delta$ entry reports variant score minus baseline score, where positive values indicate improvement over baseline and negative values indicate degradation.}
\label{tab:segment-ablation-50}
\end{table}

The experimental results are shown in Table~\ref{tab:segment-ablation-50}. The updated delta table shows a clear pattern: all ablation families are negative on average, meaning the final SAPO prompt remains the strongest configuration overall. The largest degradation comes from reverting tasks to initial wording (Avg \(\Delta=-0.1038\)), with the strongest drop on GSM8K (\(-0.2800\)); this confirms that task-level refinement is the main driver of improvements.

Removing response-format constraints also hurts on average (Avg \(\Delta=-0.0354\)), and role removal is similarly unfavorable overall (Avg \(\Delta=-0.0291\)). Although isolated positives exist (e.g., SQuADv2 for \(-\)Role, GSM8K for \(-\)RespFmt), they are not strong enough to reverse the cross-dataset trend.

Context reversion is the least harmful perturbation (Avg \(\Delta=-0.0031\)), with near-zero effects on SQuADv2 and GSM8K and mixed small effects elsewhere. This suggests that, in the current setup, context edits are comparatively stable, while tasks and output constraints are the most sensitive levers.

\subsection{Computational Costs.}
\label{app:costs}
To make algorithmic cost comparable across methods, we use a call-level proxy under a common budget (\(T\) iterations, \(K\) candidates per iteration). Let \(N_{tr}\) and \(N_{val}\) denote train and validation sizes, and let \(c_{diag}\), \(c_{refl}\), and \(c_{gen}\) denote per-iteration diagnostic/reflection/generation overhead constants. Under this proxy:
\begin{itemize}

\item APO:
\[
C_{\text{APO}}
=
T\Big(
K\,N_{tr}
+
K\,N_{val}
+
c_{diag}
+
c_{refl}
+
K\,c_{gen}
\Big)
\]

\item OPRO:
\[
C_{\text{OPRO}}
=
T\Big(
K\,N_{val}
+
K\,c_{gen}
\Big)
\]

\item EvoPrompt:
\[
C_{\text{Evo}}
=
T\Big(
K\,N_{val}
+
K\,c_{gen}
\Big)
\]

\item StraGO:
\[
C_{\text{StraGO}}
=
T\Big(
K\,N_{tr}
+
K\,N_{val}
+
c_{diag}
+
K\,c_{gen}
\Big)
\]

\item GEPA:
\[
C_{\text{GEPA}}
=
T\Big(
K\,N_{tr}
+
K\,N_{val}
+
c_{diag}
+
c_{refl}
+
K\,c_{gen}
\Big)
\]

\item SAPO:
\[
C_{\text{SAPO}}
=
T\Big(
N_{tr}
+
K\,N_{val}
+
c_{diag}
+
c_{gen}
\Big)
\]

\end{itemize}

This proxy captures dominant LLM-call complexity, while wall-clock time also depends on provider latency, batching, and parallelization.

\section{Prompt Optimization Example by SAPO}
\begin{figure*}
\centering
\includegraphics[width=0.8\linewidth]{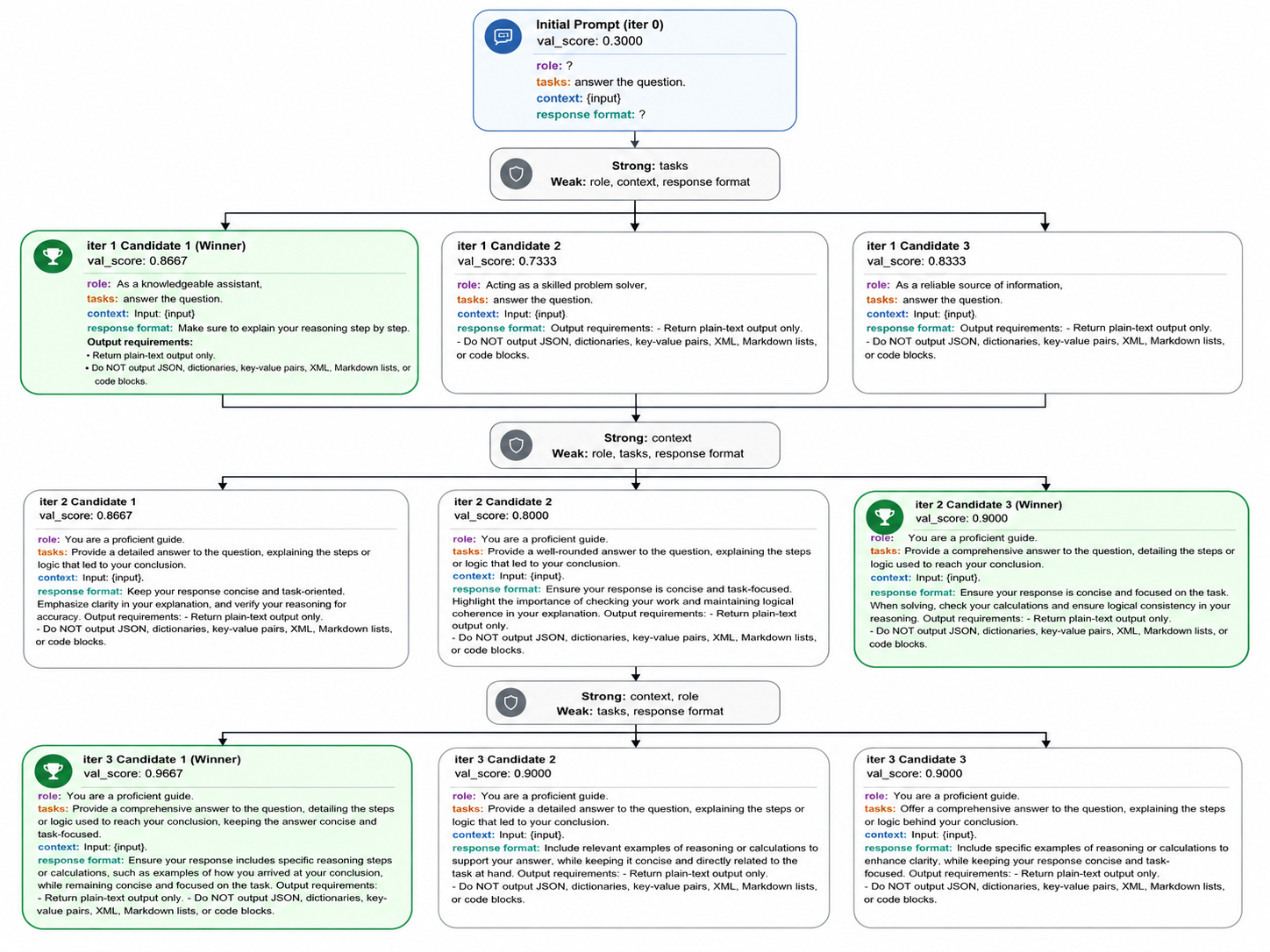}
\caption{Example of SAPO optimization on GSM8K samples. The figure traces how the LLM diagnoses and updates weak prompt segments across iterations. At each step, performance is analyzed at the segment level; missing segments are treated as weak by default. Segments identified as strong are preserved and not modified during candidate generation. The best result was obtained at iteration 3 (candidate 1), where segment-level expansion and reformulation yielded near-complete coverage of correct answers on the sample.}
\label{fig:sapo_e5_temp}
\end{figure*}

\newpage

\section{SAPO Optimized Prompts}
\label{appendix:prompt}
The figures below show, for each dataset, the initial prompt (top) and the optimized segmented prompt (bottom) used in GPT-4o-mini experiments.

\begin{figure}[h]
\centering
\includegraphics[width=\linewidth]{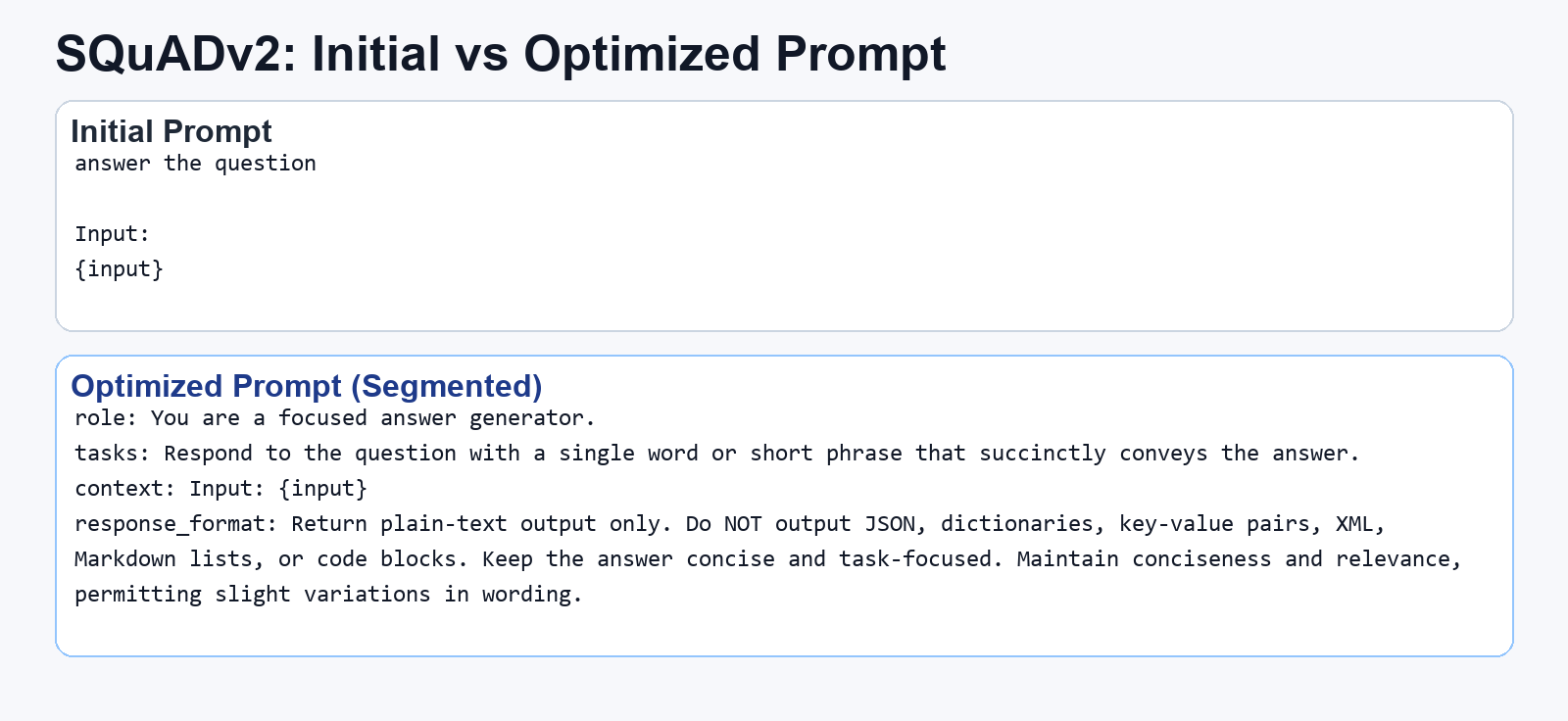}
\caption{SQuADv2 prompt pair (GPT-4o-mini): initial prompt (top) and optimized segmented prompt (bottom).}
\label{fig:prompt_pair_squadv2}
\end{figure}

\begin{figure}[h]
\centering
\includegraphics[width=\linewidth]{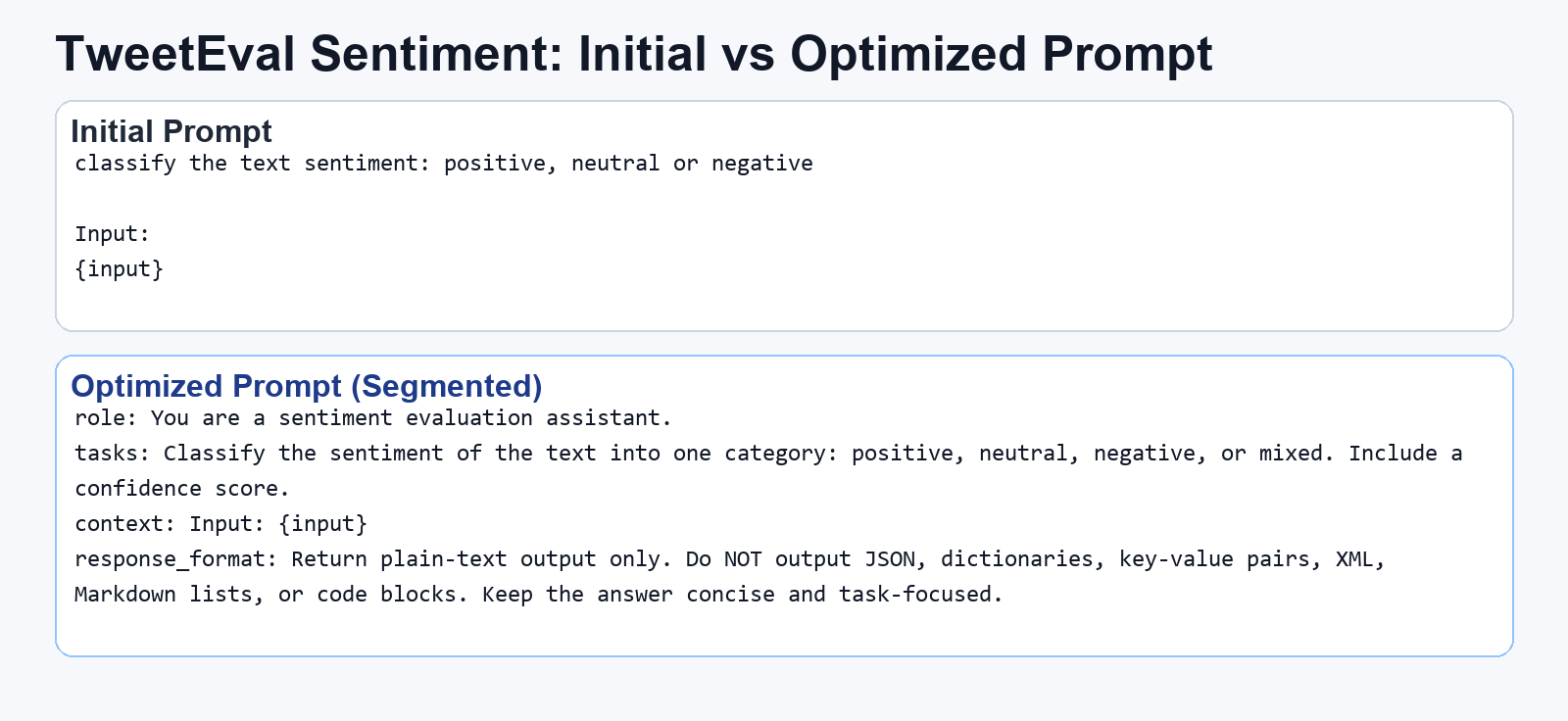}
\caption{TweetEval Sentiment prompt pair (GPT-4o-mini): initial prompt (top) and optimized segmented prompt (bottom).}
\label{fig:prompt_pair_tweeteval}
\end{figure}

\begin{figure}[h]
\centering
\includegraphics[width=\linewidth]{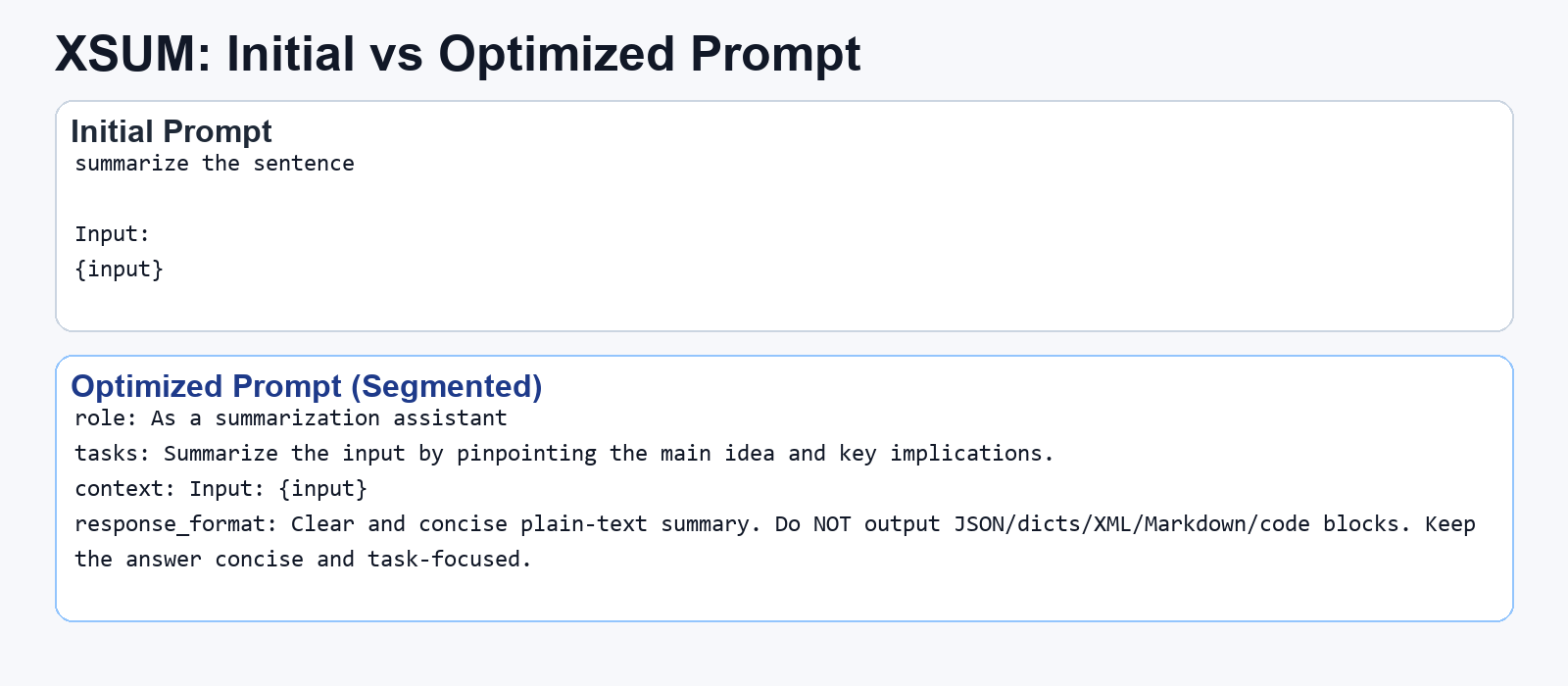}
\caption{XSUM prompt pair (GPT-4o-mini): initial prompt (top) and optimized segmented prompt (bottom).}
\label{fig:prompt_pair_xsum}
\end{figure}

\begin{figure}[h]
\centering
\includegraphics[width=\linewidth]{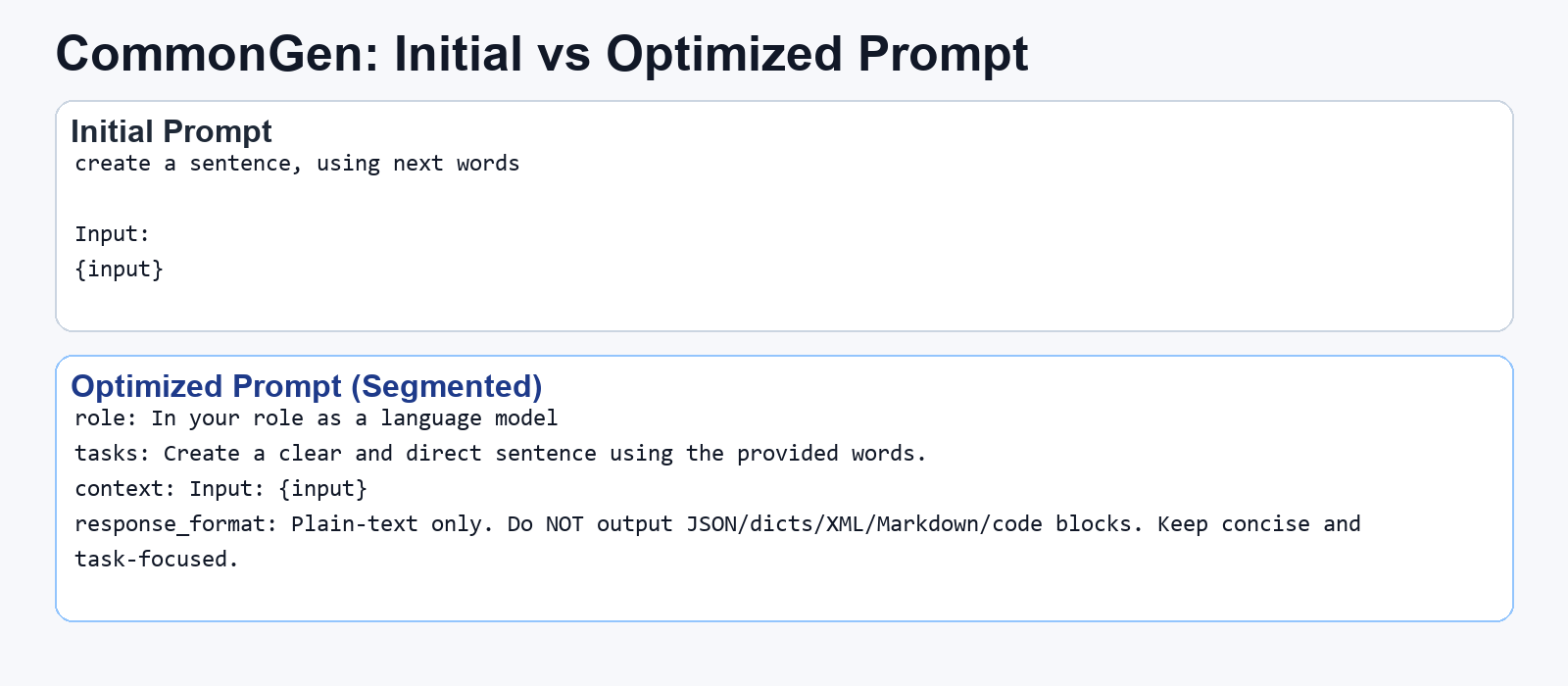}
\caption{CommonGen prompt pair (GPT-4o-mini): initial prompt (top) and optimized segmented prompt (bottom).}
\label{fig:prompt_pair_commongen}
\end{figure}

\begin{figure}[h]
\centering
\includegraphics[width=\linewidth]{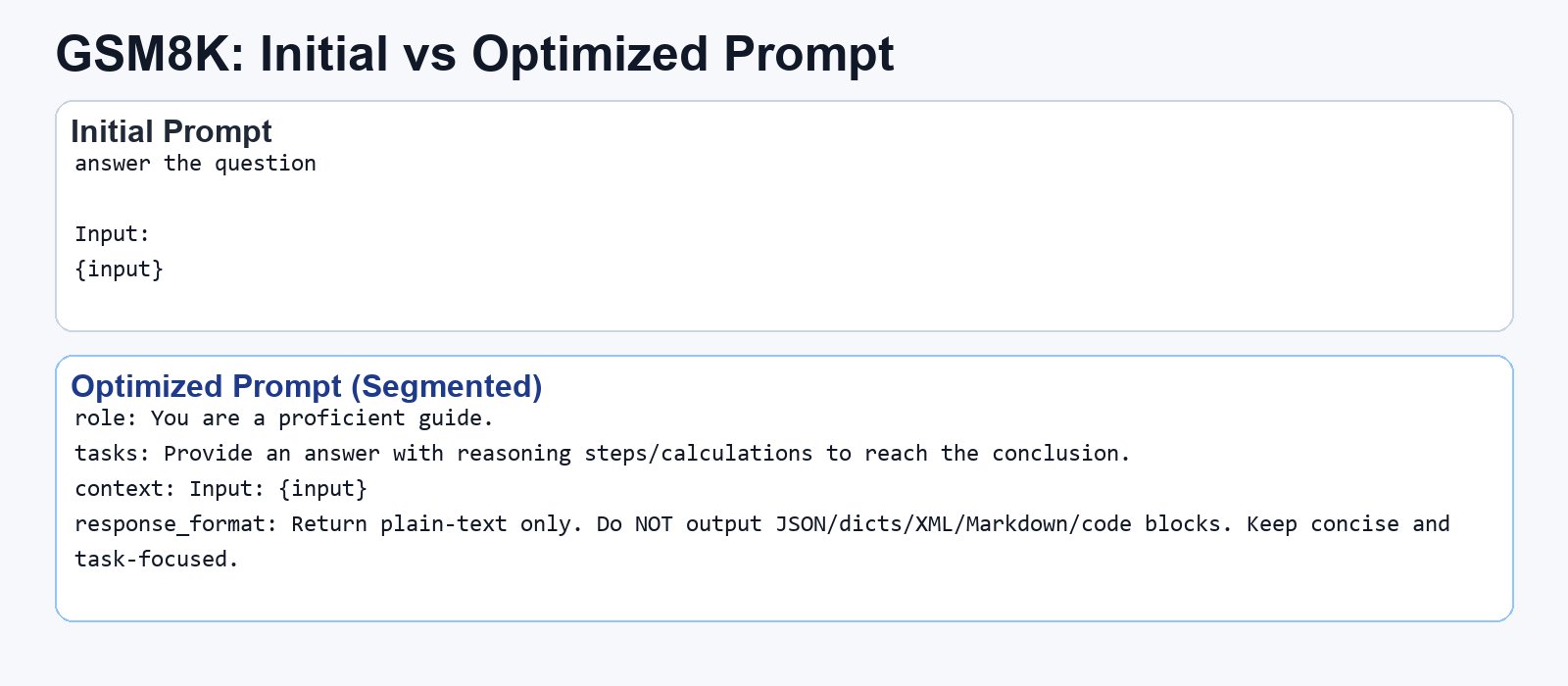}
\caption{GSM8K prompt pair (GPT-4o-mini): initial prompt (top) and optimized segmented prompt (bottom).}
\label{fig:prompt_pair_gsm8k}
\end{figure}

\begin{figure}[h]
\centering
\includegraphics[width=\linewidth]{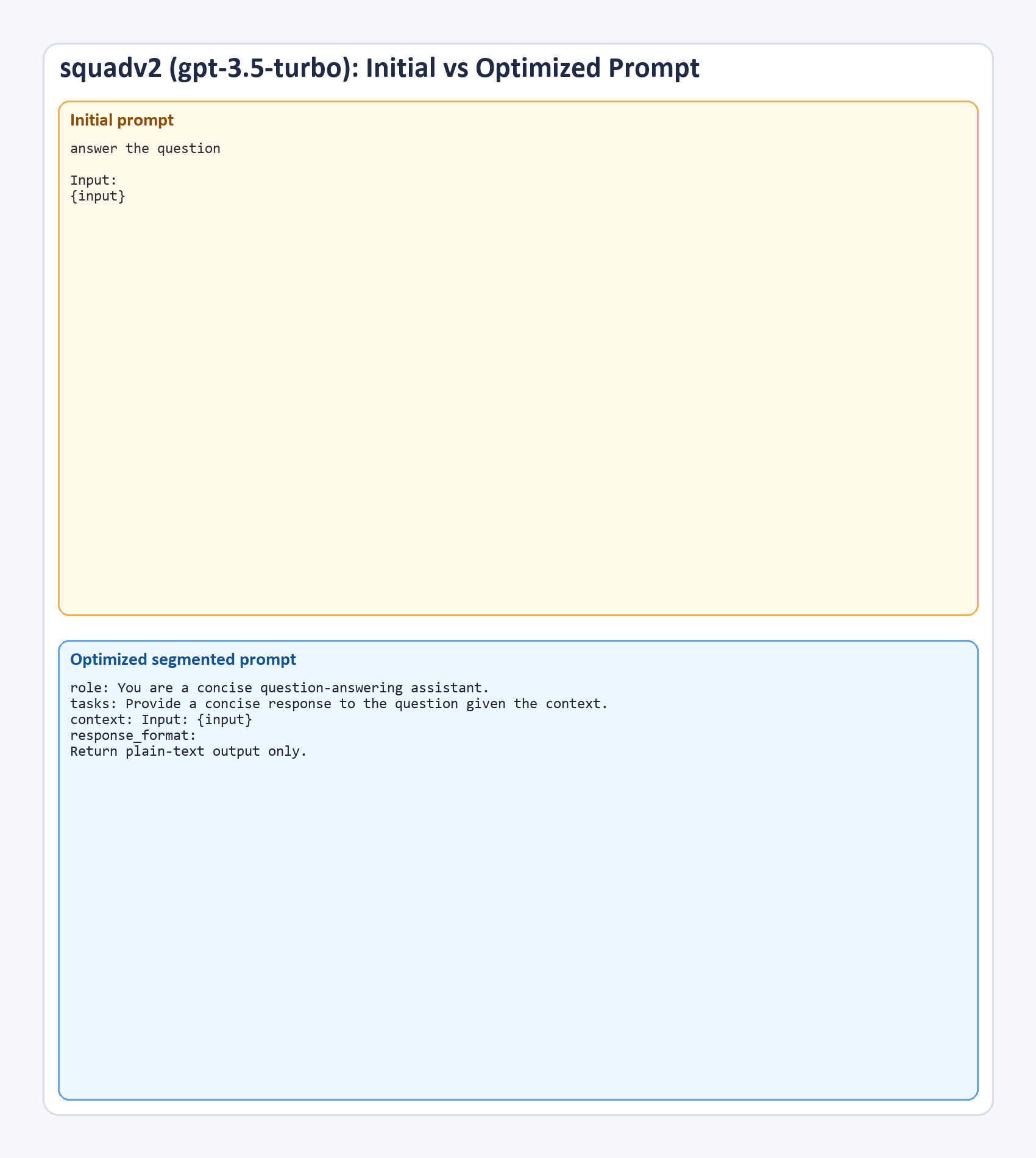}
\caption{SQuADv2 prompt pair (GPT-3.5-Turbo): initial prompt (top) and optimized segmented prompt (bottom).}
\label{fig:prompt_pair_squadv2_gpt35}
\end{figure}

\begin{figure}[h]
\centering
\includegraphics[width=\linewidth]{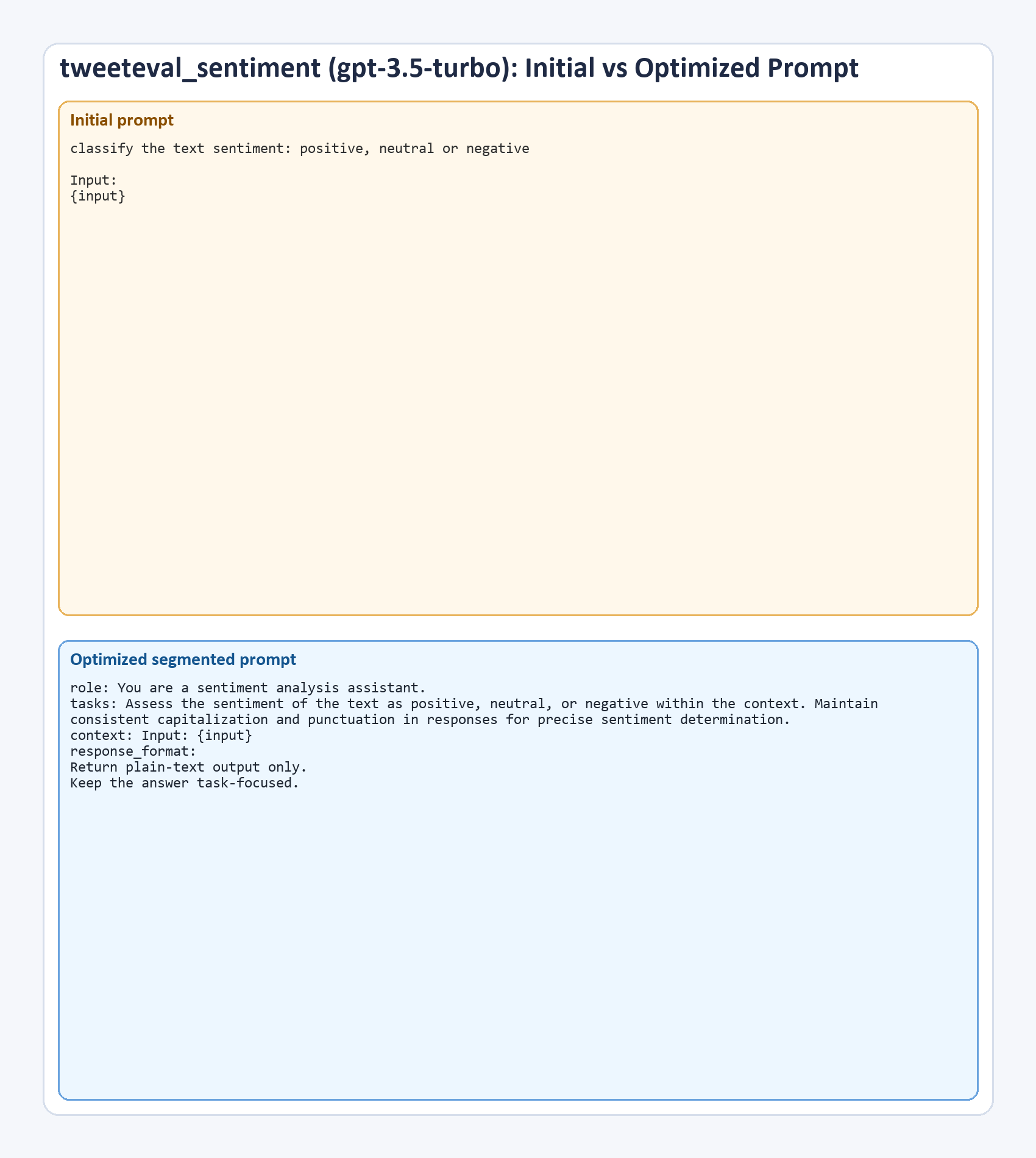}
\caption{TweetEval Sentiment prompt pair (GPT-3.5-Turbo): initial prompt (top) and optimized segmented prompt (bottom).}
\label{fig:prompt_pair_tweeteval_gpt35}
\end{figure}

\begin{figure}[h]
\centering
\includegraphics[width=\linewidth]{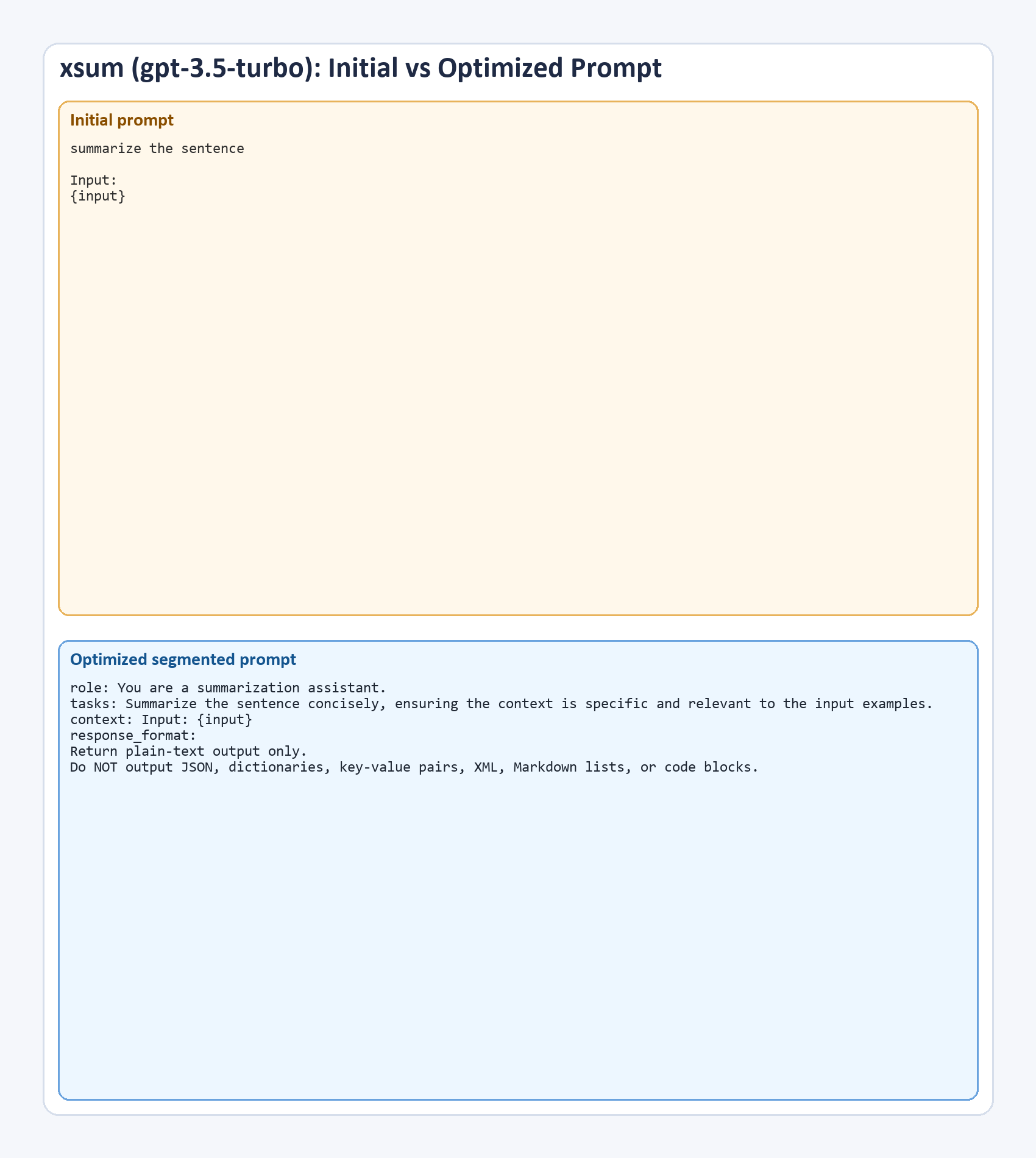}
\caption{XSUM prompt pair (GPT-3.5-Turbo): initial prompt (top) and optimized segmented prompt (bottom).}
\label{fig:prompt_pair_xsum_gpt35}
\end{figure}

\begin{figure}[h]
\centering
\includegraphics[width=\linewidth]{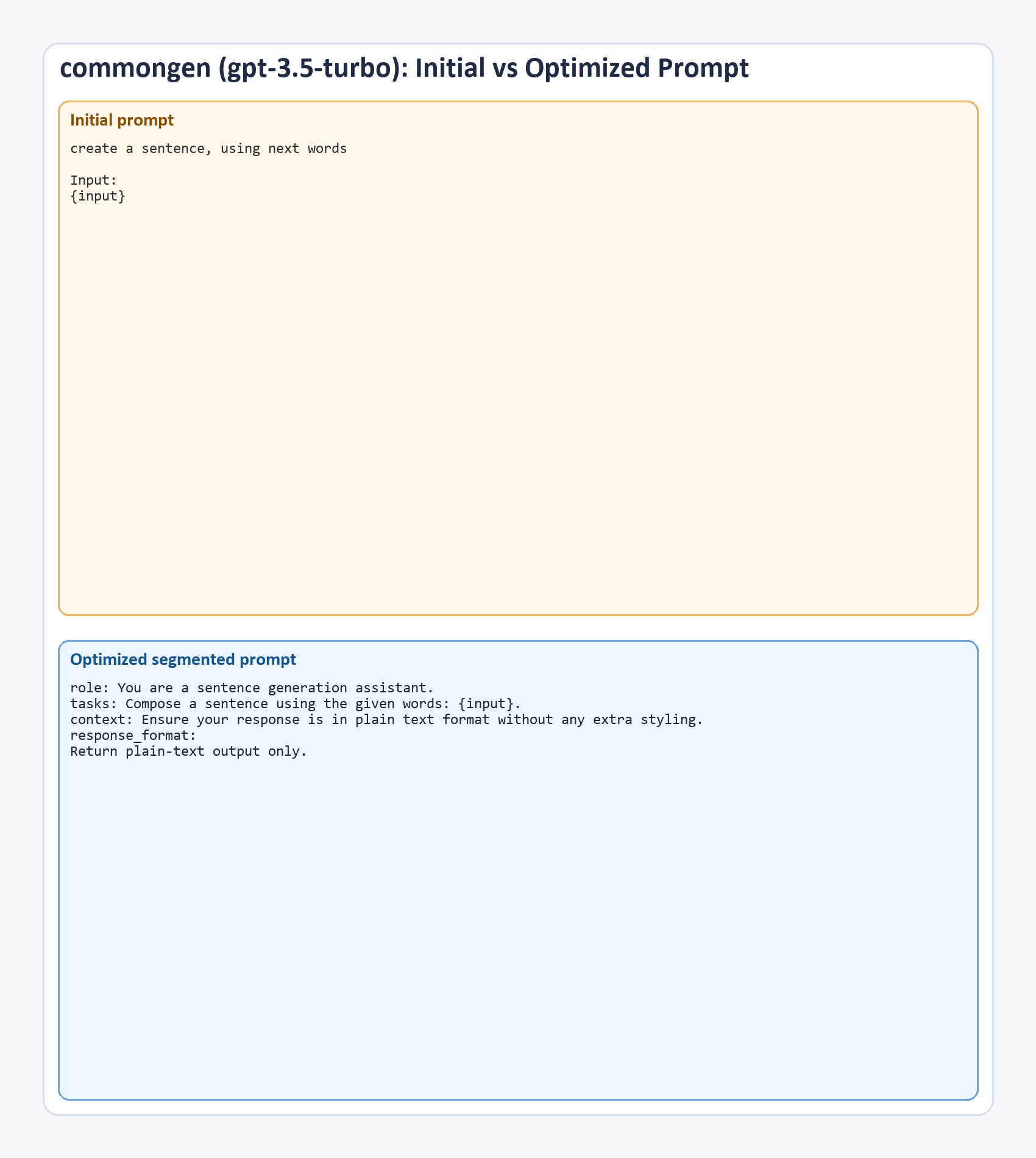}
\caption{CommonGen prompt pair (GPT-3.5-Turbo): initial prompt (top) and optimized segmented prompt (bottom).}
\label{fig:prompt_pair_commongen_gpt35}
\end{figure}

\begin{figure}[h]
\centering
\includegraphics[width=\linewidth]{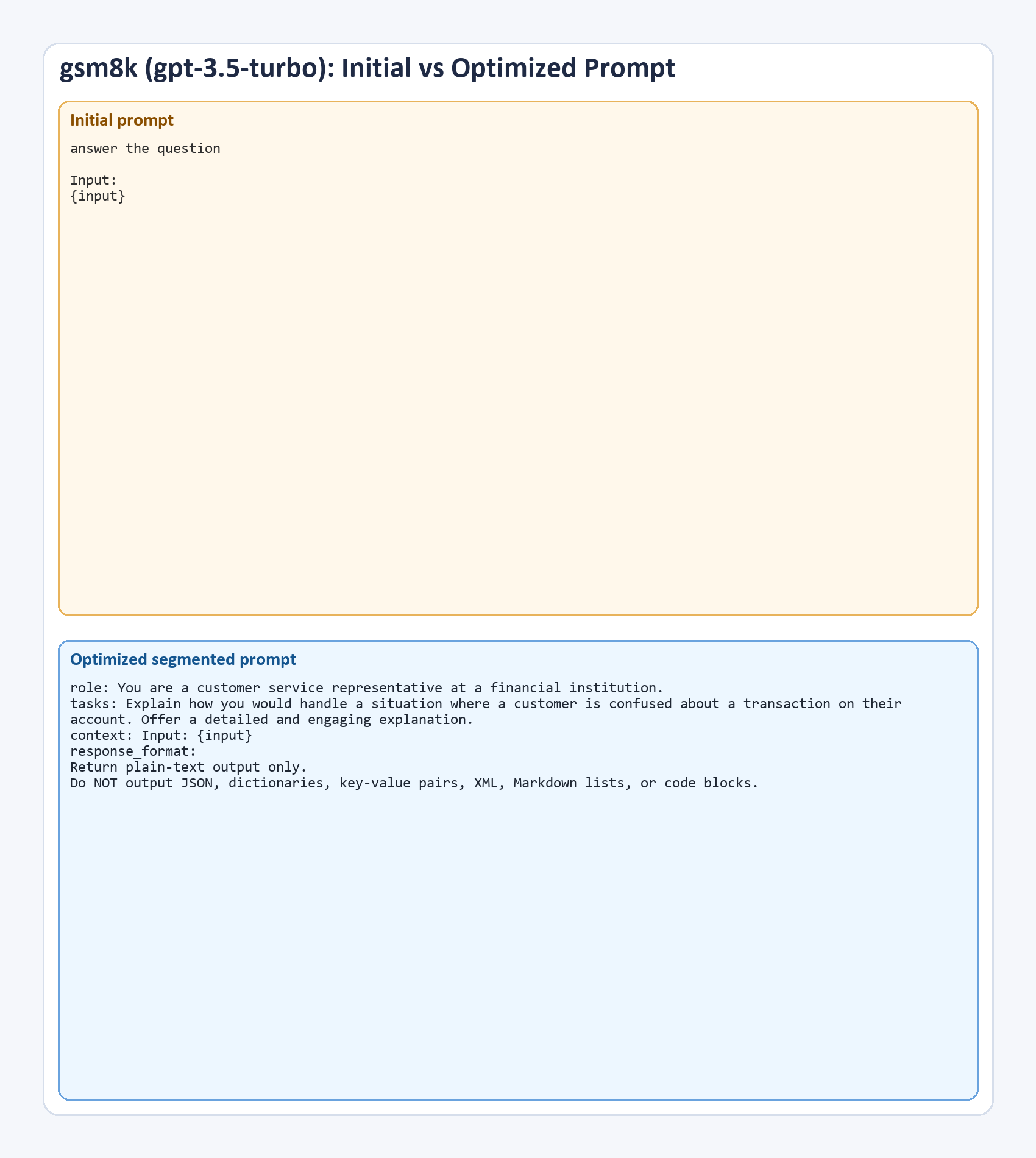}
\caption{GSM8K prompt pair (GPT-3.5-Turbo): initial prompt (top) and optimized segmented prompt (bottom).}
\label{fig:prompt_pair_gsm8k_gpt35}
\end{figure}

\FloatBarrier
\clearpage

\section{Meta-prompts}
\label{appendix:meta-prompt}

\begin{figure}[h]
\centering
\includegraphics[width=1\linewidth]{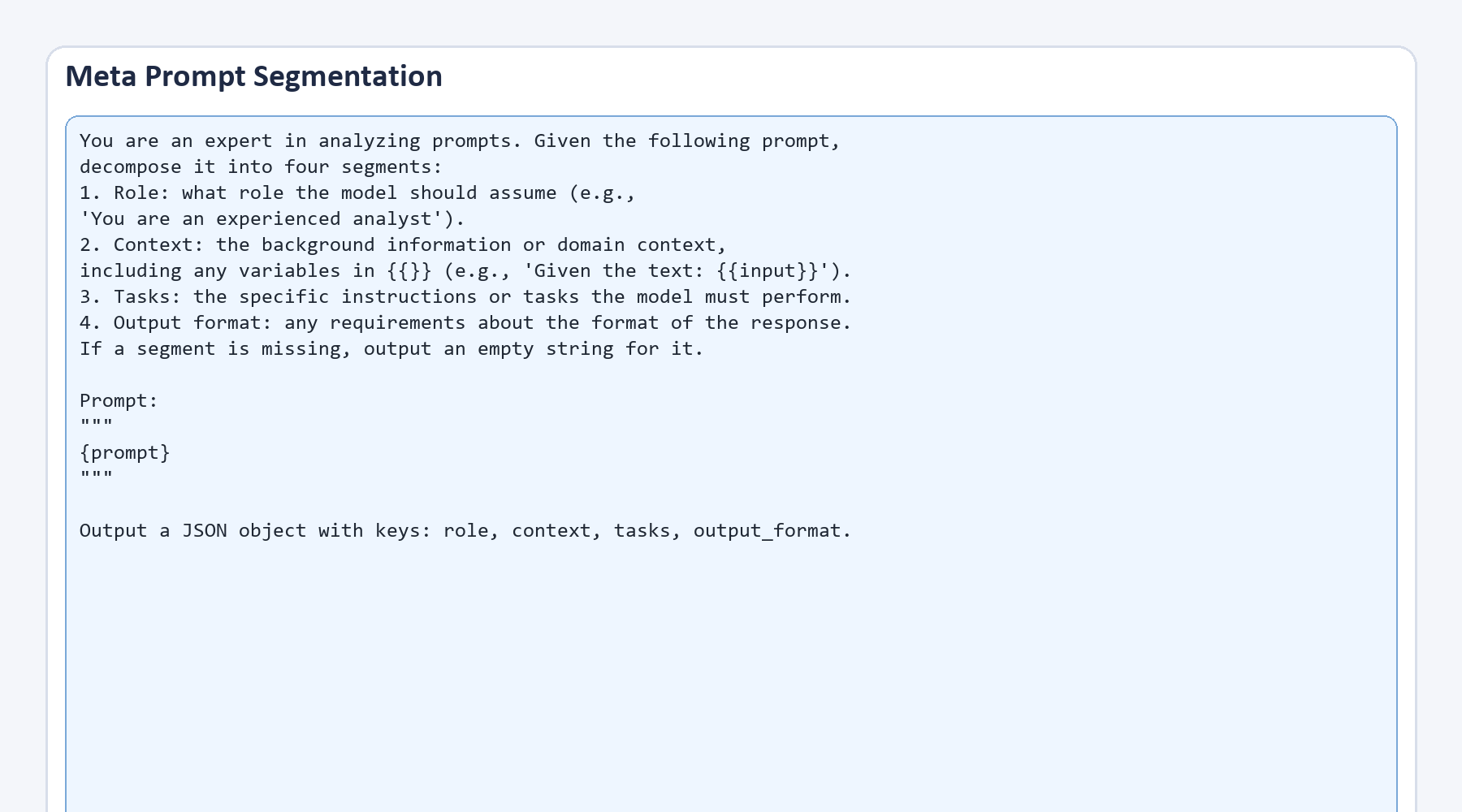}
\caption{Segmentation meta-prompt used by SAPO to decompose a prompt into role, context, tasks, and output format.}
\label{fig:meta_prompt_segmentation}
\end{figure}

\begin{figure}[h]
\centering
\includegraphics[width=1\linewidth]{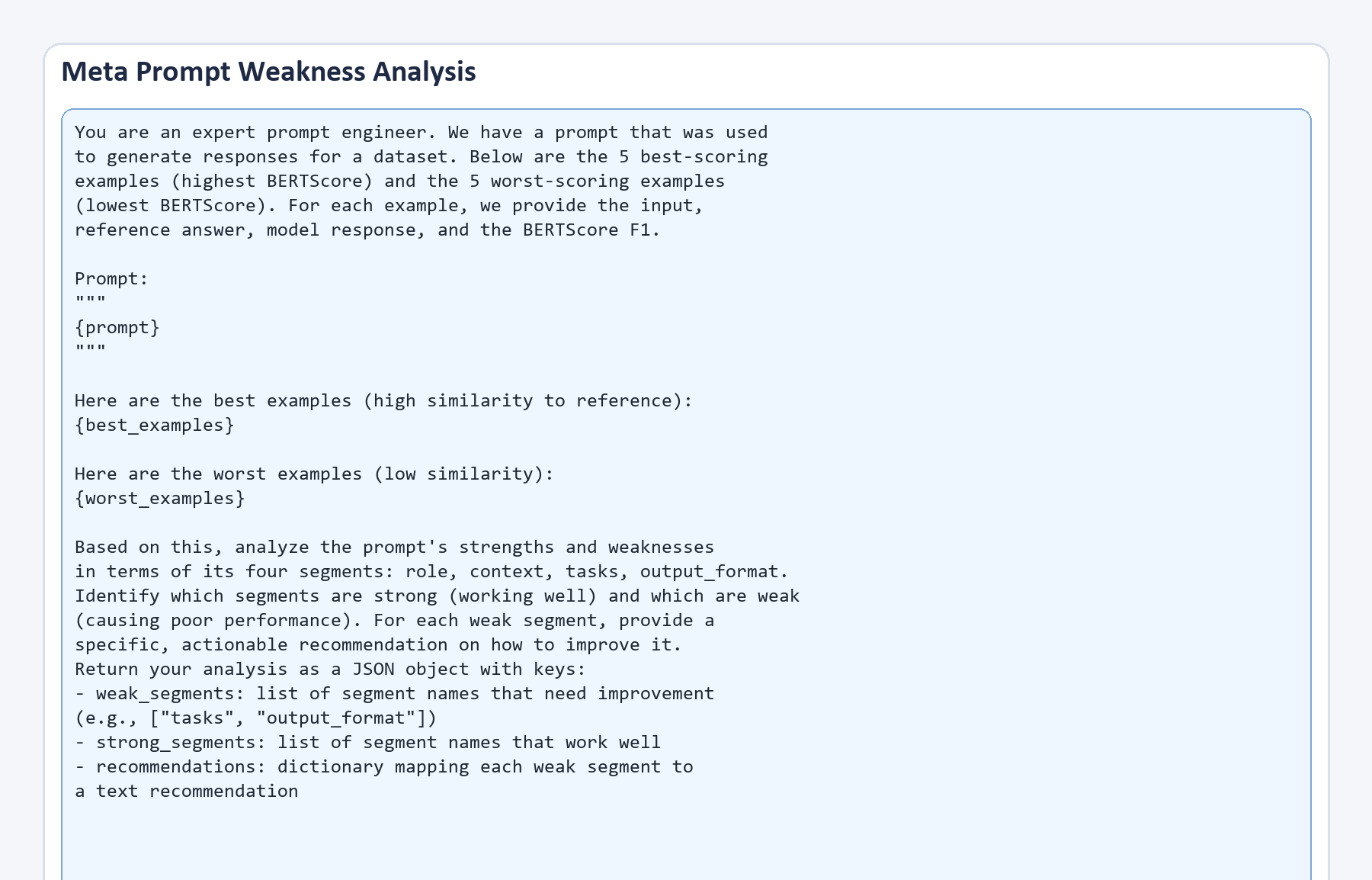}
\caption{Weakness-analysis meta-prompt: SAPO contrasts top/bottom evidence and outputs weak segments, strong segments, and actionable recommendations.}
\label{fig:meta_prompt_weakness}
\end{figure}

\begin{figure}[h]
\centering
\includegraphics[width=1\linewidth]{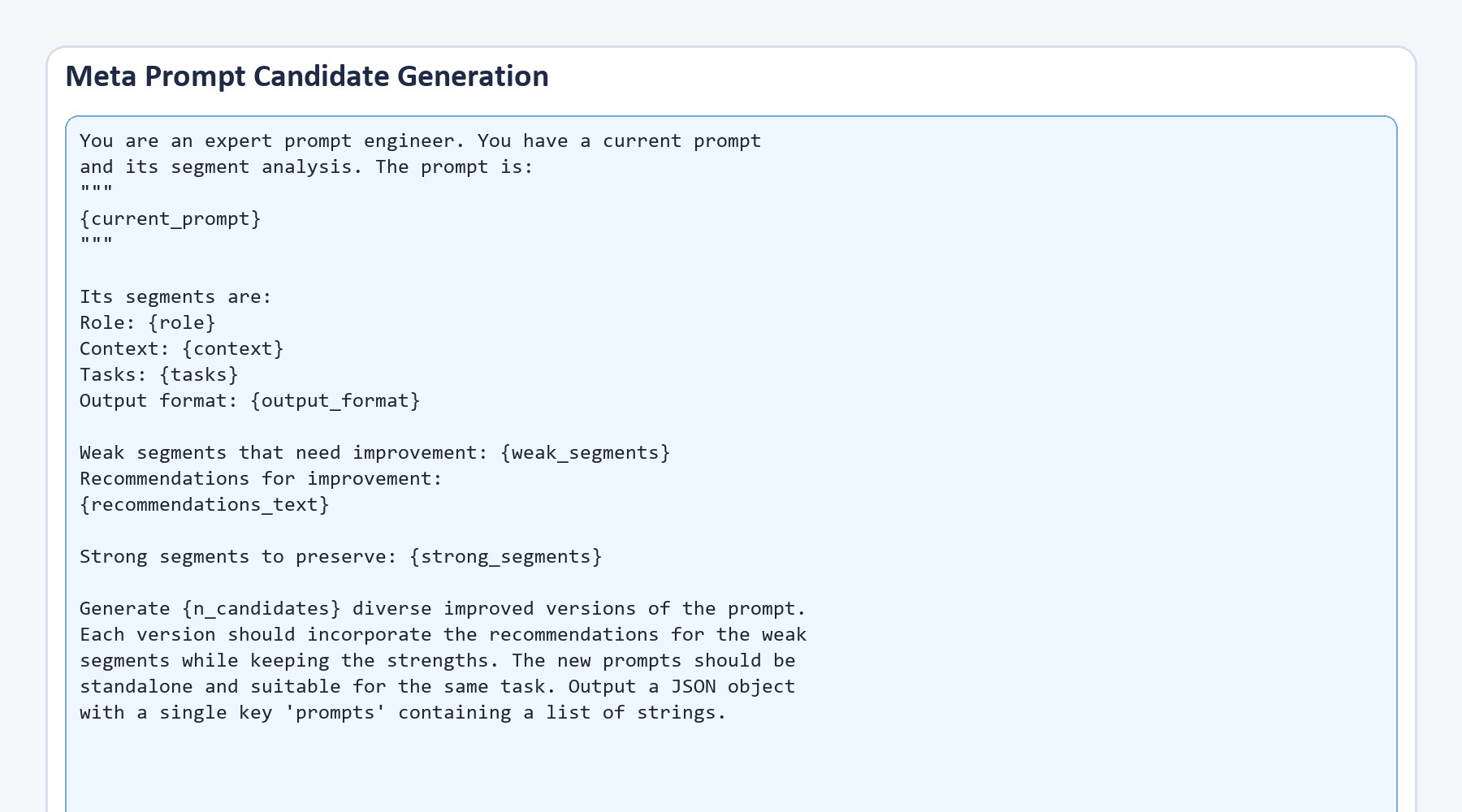}
\caption{Candidate-generation meta-prompt used to synthesize constrained prompt variants that preserve strong segments and revise weak ones.}
\label{fig:meta_prompt_candidate}
\end{figure}

\end{document}